\documentclass[10pt,twocolumn,letterpaper]{article}

\usepackage{iccv}
\usepackage{times}
\usepackage{epsfig}
\usepackage{graphicx}
\usepackage{amsmath}
\usepackage{amssymb}
\usepackage{xcolor}
\usepackage{adjustbox}
\usepackage{color, colortbl}
\definecolor{citecolor}{HTML}{2980b9}
\definecolor{linkcolor}{HTML}{c0392b}

\usepackage{booktabs}
\usepackage{multirow}
\usepackage{makecell}
\newcommand\blfootnote[1]{%
  \begingroup
  \renewcommand\thefootnote{}\footnote{#1}%
  \addtocounter{footnote}{-1}%
  \endgroup
}
\usepackage[accsupp]{axessibility}

\usepackage[hidelinks,pagebackref=true,breaklinks=true,colorlinks,bookmarks=false,citecolor=citecolor,letterpaper=true,linkcolor=linkcolor]{hyperref} 

\iccvfinalcopy 

\def\iccvPaperID{1371} 

\ificcvfinal\fi

\begin{document}

\title{MonoDETR: Depth-guided Transformer for Monocular 3D Object Detection}

\author{Renrui Zhang$^{1,2}$, Han Qiu$^{2}$, Tai Wang$^{1,2}$, Ziyu Guo$^{2}$, Yiwen Tang$^{2}$, Xuanzhuo Xu$^{2}$\\Ziteng Cui$^{2}$, Yu Qiao$^{2}$, Hongsheng Li$^{\dagger1,2,3}$, Peng Gao$^{\dagger2}$\vspace{0.3cm}\\
  $^1$CUHK MMLab\quad 
  $^2$Shanghai Artificial Intelligence Laboratory\\
  $^3$Centre for Perceptual and Interactive Intelligence (CPII)\vspace{0.2cm}\\
\texttt{\{zhangrenrui, wangtai, gaopeng\}@pjlab.org.cn},\quad
\texttt{hsli@ee.cuhk.edu.hk}
}

\maketitle
\ificcvfinal\thispagestyle{empty}\fi

\blfootnote{$\dagger$ Corresponding author}
\begin{abstract}
   Monocular 3D object detection has long been a challenging task in autonomous driving. Most existing methods follow conventional 2D detectors to first localize object centers, and then predict 3D attributes by neighboring features. However, only using local visual features is insufficient to understand the scene-level 3D spatial structures and ignores the long-range inter-object depth relations. 
    In this paper, we introduce the first DETR framework for \textbf{Mono}cular \textbf{DE}tection with a depth-guided \textbf{TR}ansformer, named \textbf{MonoDETR}. We modify the vanilla transformer to be depth-aware and guide the whole detection process by contextual depth cues. Specifically, concurrent to the visual encoder that captures object appearances, we introduce to predict a foreground depth map, and specialize a depth encoder to extract non-local depth embeddings. Then, we formulate 3D object candidates as learnable queries and propose a depth-guided decoder to conduct object-scene depth interactions.
    In this way, each object query estimates its 3D attributes adaptively from the depth-guided regions on the image and is no longer constrained to local visual features.
    On KITTI benchmark with monocular images as input, MonoDETR achieves \textit{state-of-the-art} performance and requires no extra dense depth annotations. Besides, our depth-guided modules can also be plug-and-play to enhance multi-view 3D object detectors on nuScenes dataset, demonstrating our superior generalization capacity. Code is available at \url{https://github.com/ZrrSkywalker/MonoDETR}.
\end{abstract}

\section{Introduction}

With a wide range of applications in autonomous driving, 3D object detection is more challenging than its 2D counterparts, due to the complex spatial circumstances. 
Compared to methods based on LiDAR~\cite{VoxelNet,PointPillars,PointRCNN,CenterPoint} and multi-view images~\cite{detr3d,li2022bevformer,liu2022petr,huang2021bevdet}, 3D object detection from monocular (single-view) images~\cite{3DOP,M3D-RPN,fcos3d} is of most difficulty, which generally does not rely on depth measurements or multi-view geometry. The detection accuracy thus severely suffers from the ill-posed depth estimation, leading to inferior performance.

\begin{figure}[t]
  \centering
    \includegraphics[width=0.45\textwidth]{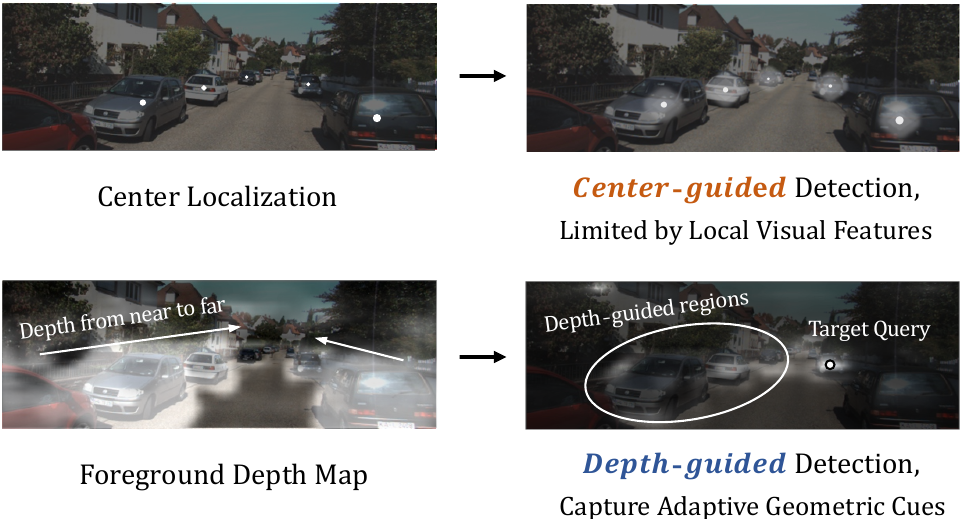}
    \vspace{0.25cm}
   \caption{\textbf{Center-guided (Top) and Depth-guided Paradigms (Bottom) for monocular 3D object detection.} Existing center-guided methods predict 3D attributes from local visual features around the centers, while our MonoDETR guides the detection by a predicted foreground depth map and adaptively aggregates features in global context. The lower right figure visualizes the attention map of the target query in the depth cross-attention layer.}
    \label{fig1}
\end{figure}

Except for leveraging pseudo 3D representations~\cite{PseudoLiDAR,PseudoLiDAR++,patchnet,CaDDN}, standard monocular 3D detection methods~\cite{monodle,MonoFlex,monogeo,gupnet} follow the pipeline of traditional 2D object detection~\cite{ren2015faster,RetinaNet,tian2019fcos,CenterNet}. They first localize objects by detecting the projected centers on the image, and then aggregate the neighboring visual features for 3D property prediction, as illustrated in Figure~\ref{fig1} (Top). Although it is conceptually straightforward, such center-guided methods are limited by the local appearances without long-range context, and fail to capture implicit geometric cues from 2D images, e.g., depth guidance, which are critical to detect objects in 3D space.

\begin{figure*}[t!]
  \centering
    \includegraphics[width=0.97\textwidth]{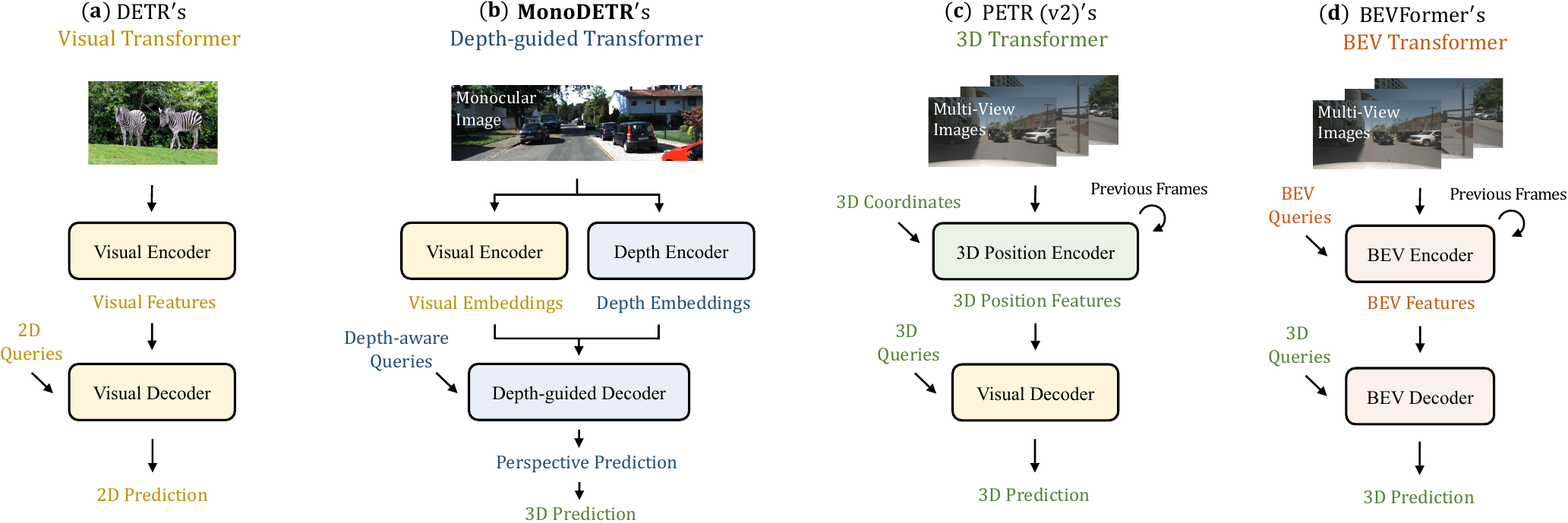}
    \vspace{0.1cm}
   \caption{\textbf{Comparison of DETR-based methods for camera-based 3D object detection.} We utilize yellow, blue, green, and red to respectively denote the feature or prediction space related with 2D, depth, 3D, and BEV. Different from other methods, our MonoDETR leverages depth cues to guide 3D object detection from monocular images.}
    \label{fig2}
\end{figure*}

To tackle this issue, we propose MonoDETR, which presents a novel depth-guided 3D detection scheme in Figure~\ref{fig1} (Bottom). Compared to DETR~\cite{DETR} in 2D detection, the transformer in MonoDETR is equipped with depth-guided modules to better capture contextual depth cues, serving as \textit{\textbf{the first}} DETR model for monocular 3D object detection, as shown in Figure~\ref{fig2} (a) and (b). It consists of: two parallel encoders for visual and depth representation learning, and a decoder for adaptive depth-guided detection.

Specifically, after the feature backbone, we first utilize a lightweight depth predictor to acquire the depth features of the input image. To inject effective depth cues, a foreground depth map is predicted on top, and supervised only by discrete depth labels of objects, which requires no dense depth annotations during training. 
Then, we apply the parallel encoders to respectively generate non-local depth and visual embeddings, which represent the input image from two aspects: depth geometry and visual appearance. 
On top of that, a set of object queries is fed into the depth-guided decoder, and conducts adaptive feature aggregation from the two embeddings. 
Via a proposed depth cross-attention layer, the queries can capture geometric cues from the depth-guided regions on the image, and explore inter-object depth relations.
In this way, the 3D attribute prediction can be guided by informative depth hints, no longer constrained by the limited visual features around centers.

As an end-to-end transformer-based network, MonoDETR is free from non-maximum suppression (NMS) or rule-based label assignment. We only utilize the object-wise labels for supervision without using auxiliary data, such as dense depth maps or LiDAR. Taking monocular images as input, MonoDETR achieves \textit{state-of-the-art} performance among existing center-guided methods, and surpasses the second-best by +2.53\%, +1.08\%, and +0.85\% for three-level difficulties on KITTI~\cite{kitti} \textit{test} set.

Besides single-view images, the depth-guided modules in MonoDETR can also be extended as a plug-and-play module for multi-view 3D detection on nuScenes~\cite{nuScenes} dataset. By providing multi-view depth cues, our method can not only improve the end-to-end detection performance of PETRv2~\cite{liu2022petrv2} by +1.2\% NDS, but also benefit the BEV representation learning in BEVFormer~\cite{li2022bevformer} by +0.9\% NDS. This further demonstrates the effectiveness and generalizability of our proposed depth guidance. 

We summarize the contributions of our paper as follows:
\begin{itemize}
    \item We propose MonoDETR, a depth-guided framework to capture scene-level geometries and inter-object depth relations for monocular 3D object detection.\vspace{-0.1cm}
    \item We introduce a foreground depth map for object-wise depth supervision, and a depth cross-attention layer for adaptive depth features interaction.\vspace{-0.1cm}
    \item MonoDETR achieves leading results on monocular KITTI benchmark, and can also be generalized to enhance multi-view detection on nuScenes benchmark.
\end{itemize}

\section{Related Work}

Existing methods for camera-based 3D object detection can be categorized as two groups according to the input number of views: monocular (single-view) and multi-view methods. Monocular detectors only take as input the front-view images and solve a more challenging task from insufficient 2D signals. Multi-view detectors simultaneously encode images of surrounding scenes and can leverage cross-view dependence to understand the 3D space.
\vspace{-0.3cm}

\paragraph{Monocular (Single-view) 3D Object Detection.}
Most previous monocular detectors adopt center-guided pipelines following conventional 2D detectors~\cite{ren2015faster,tian2019fcos,CenterNet}. As early works, Deep3DBox~\cite{Deep3DBox} introduces discretized representation with perspective constraints, and M3D-RPN~\cite{M3D-RPN} designs a depth-aware convolution for better 3D region proposals. With very few handcrafted modules, SMOKE~\cite{smoke} and FCOS3D~\cite{fcos3d} propose concise architectures for efficient one-stage detection, while MonoDLE~\cite{monodle} and PGD~\cite{PGD} analyze depth errors on top with improved performance. 
To supplement the limited 3D cues, additional data are utilized for assistance: dense depth annotations~\cite{patchnet,D4LCN, ddpm,park2021pseudo}, CAD models~\cite{AutoShape}, and LiDAR~\cite{monorun, CaDDN, monodtr}. Some recent methods introduce complicated geometric priors into the networks: adjacent object pairs~\cite{MonoPair}, 2D-3D keypoints~\cite{RTM3D}, and uncertainty-related depth~\cite{MonoFlex,gupnet}. Despite this, the center-guided methods are still limited by local visual features without scene-level spatial cues. In contrast, MonoDETR discards the center localization step and conducts adaptive feature aggregation via a depth-guided transformer. MonoDETR requires no additional annotations and contains minimal 2D-3D geometric priors.
\vspace{-0.3cm}

\begin{table}[t]
\centering
\small
\vspace{0.1cm}
\begin{adjustbox}{width=\linewidth}
	\begin{tabular}{lccccc}
	\toprule
	  \makecell[l]{Method} &\makecell[c]{DETR\\-based} &\makecell[c]{Extra\\Data} &\makecell[c]{Guided\\by} &\makecell[c]{Object\\Query} &\makecell[c]{Feat.\\Aggre.} \\
   \cmidrule(lr){1-1} \cmidrule(lr){2-6}
        \color{gray}{\textit{Mutli-view Methods}}\\
        DETR3D~\cite{detr3d} &\checkmark &- &Visual &3D &Local\\
        PETR (v2)~\cite{liu2022petr} &\checkmark &Temporal &Visual &3D &Global \\
        BEVFormer~\cite{li2022bevformer} &\checkmark &Temporal &Visual  &BEV, 3D &Global\\
        \cmidrule(lr){1-6}
        \color{gray}{\textit{Monocular Methods}}\\
        MonoDTR~\cite{monodtr} &$\times$ &LiDAR &Center &$\times$ &Local \\
        MonoDETR &\checkmark &- &\textbf{Depth} &\textbf{Depth} &Global\\
	\bottomrule
	\end{tabular}
\end{adjustbox}
\vspace{0.1cm}
\caption{\textbf{Comparison of DETR-based methods for camera-based 3D object detection.} Our MonoDETR is uniquely guided by depth cues with depth-aware object queries.}
\vspace{0.1cm}
 \label{t1}
\end{table}

\paragraph{Multi-view 3D Object Detection.}
For jointly extracting features from surrounding views, DETR3D~\cite{detr3d} firstly utilizes a set of 3D object queries and back-projects them onto multi-view images for feature aggregation. PETR series~\cite{liu2022petr,liu2022petrv2} further proposes to generate 3D position features without unstable projection and explores the advantage of temporal information from previous frames. From another point of view, BEVDet~\cite{huang2021bevdet,huang2022bevdet4d} follows~\cite{philion2020lift} to lift 2D images into a unified Bird’s-Eye-View (BEV) representation and appends BEV-based heads~\cite{CenterPoint} for detection. BEVFormer~\cite{li2022bevformer} instead generates BEV features via a set of learnable BEV queries, and introduces a spatiotemporal BEV transformer for visual features aggregation. Follow-up works also introduce cross-modal distillation~\cite{huang2023tig,chen2022bevdistill} and masked image modeling~\cite{liu2023towards,chen2023pimae} for improved performance. Different from the above methods for multi-view input, MonoDETR targets monocular images and extracts depth guidance to capture more geometric cues. Our depth-guided modules can also be generalized to surrounding views as a plug-and-play module to enhance multi-view detectors.
\vspace{-0.3cm}

\paragraph{Comparison of DETR-based Methods.}
DETR~\cite{carion2020end} and its follow-up works~\cite{gao2021fast,zhu2020deformable,zheng2020end,cond-detr} have attained great success on 2D object detection without NMS or anchors. Inspired by this, some efforts have transferred DETR into camera-based 3D object detection. We specifically compare our MonoDETR with existing DETR-based 3D object detectors in Figure~\ref{fig2} and Table~\ref{t1}. 
\textbf{(1) MonoDTR~\cite{monodtr}.} Also as a single-view detector, MonoDTR utilizes transformers~\cite{vaswani2017attention} to incorporate depth features with visual representations. However, MonoDTR is \textit{not a DETR-based method}, and still adopts the traditional center-guided paradigm, which localizes objects by their centers and only aggregate local features. MonoDTR contains no object queries for global feature aggregation, and follows YOLOv3~\cite{redmon2018yolov3} to adopt complicated NMS post-processing with pre-defined anchors. 
\textbf{(2) DETR3D~\cite{detr3d} and PETR (v2)~\cite{liu2022petr,liu2022petrv2}} (Figure~\ref{fig2} (c)) are multi-view methods and follow the DETR detection pipeline. In contrast, they contain no transformer-based encoders (visual or depth), and detect objects by 3D object queries without the perspective transformation. Importantly, they are only guided by visual features and explores no depth cues from the input images.
\textbf{(3) BEVFormer~\cite{li2022bevformer}} (Figure~\ref{fig2} (d)) firstly utilizes a BEV transformer to lift multi-view images into BEV representations, and then conducts DETR-based detection within the BEV space.
Different from all aforementioned methods, MonoDETR introduces a unique depth-guided transformer that guides the 3D detection by geometric depth cues, which can generalize well to both monocular and multi-view inputs.

\begin{figure}[t]
  \centering
  \vspace{0.05cm}
    \includegraphics[width=0.48\textwidth]{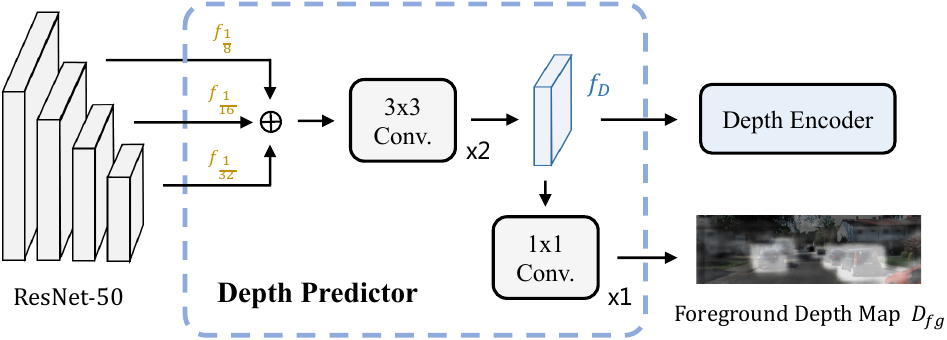}
    \vspace{0.1cm}
   \caption{\textbf{The lightweight depth predictor.} We utilize the depth predictor to predict the depth features and foreground depth map, which only contains discrete object-wise depth values.}
    \label{fig3}
    \vspace{0.1cm}
\end{figure}

\begin{figure*}[t]
  \centering
    \includegraphics[width=0.8\textwidth]{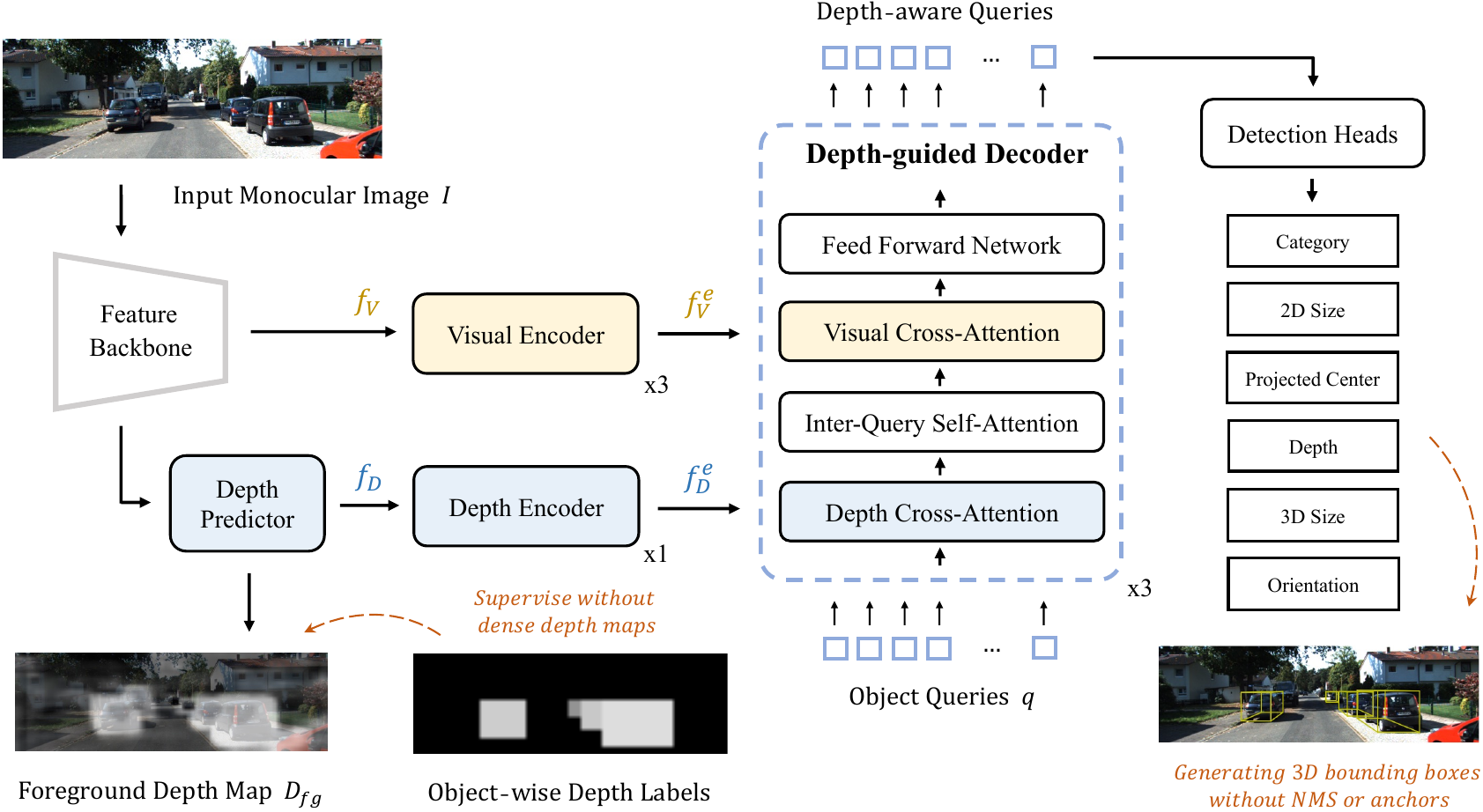}
    \vspace{0.15cm}
   \caption{\textbf{Overall pipeline of MonoDETR.} We first acquire the visual and depth features of the input image and utilize two parallel encoders for non-local encoding. Then, we propose a depth-guided decoder to adaptively aggregate scene-level features in global context.}
   \vspace{0.1cm}
    \label{fig:architecture}
\end{figure*}

\section{Method}
\label{sec:method}
The overall framework of MonoDETR is shown in Figure~\ref{fig:architecture}. We first illustrate the concurrent visual and depth feature extraction in Section~\ref{s3.1}, and detail our depth-guided transformer for aggregating appearance and geometric cues in Section~\ref{s3.2}. Then, we introduce the attribute prediction and loss functions of MonoDETR in Section~\ref{s3.3}. Finally, we illustrate how to plug our depth-guided transformer into existing multi-view object detectors in Section~\ref{s3.4}.

\subsection{Feature Extraction}

\label{s3.1}
Taking as input a monocular (single-view) image, our framework utilizes a feature backbone, e.g., ResNet-50~\cite{resnet}, and a lightweight depth predictor to generate its visual and depth features, respectively.
\vspace{-0.3cm}

\paragraph{Visual Features.}
Given the image $I \in \mathbb{R}^{H\times W\times 3}$, where $H$ and $W$ denote its height and width, we obtain its multi-scale feature maps, $f_{\frac{1}{8}}$, $f_{\frac{1}{16}}$, and $f_{\frac{1}{32}}$, from the last three stages of ResNet-50. Their downsample ratios to the original size are $\frac{1}{8}$, $\frac{1}{16}$ and $\frac{1}{32}$. We regard the highest-level $f_{\frac{1}{32}}\in \mathbb{R}^{\frac{H}{32}\times \frac{W}{32}\times C}$ with sufficient semantics as the visual features $f_V$ of the input image.
\vspace{-0.25cm}

\paragraph{Depth Features.}
We obtain the depth features from the image by a lightweight depth predictor, as shown in Figure~\ref{fig3}.
We first unify the sizes of three-level features to the same $\frac{1}{16}$ downsample ratio via bilinear pooling, and fuse them by element-wise addition. In this way, we can integrate multi-scale visual appearances and also preserve fine-grained patterns for objects of small sizes. Then, we apply two 3$\times$3 convolutional layers to obtain the depth features $f_D \in \mathbb{R}^{\frac{H}{16} \times \frac{W}{16}\times C}$ for the input image. 
\vspace{-0.25cm}

\paragraph{Foreground Depth Map.}
To incorporate effective depth information into the depth features, we predict a foreground depth map $D_{fg} \in \mathbb{R}^{\frac{H}{16}\times \frac{W}{16}\times (k+1)}$ on top of $f_D$ via a 1$\times$1 convolutional layer. We supervise the depth map only by discrete object-wise depth labels, without extra dense depth annotations. The pixels within the same 2D bounding box are assigned with the same depth label of the corresponding object. For pixels within multiple boxes, we select the depth label of the object that is nearest to the camera, which accords with the visual appearance of the image. Here, We discretize the depth into $k+1$ bins~\cite{CaDDN}, where the first ordinal $k$ bins denote foreground depth and the last one denotes the background. We adopt linear-increasing discretization (LID), since the larger depth estimation errors of farther objects can be suppressed with a wider categorization interval. We limit the foreground depth values within $[d_{min}, d_{max}]$, and set both the first interval length and LID's common difference as $\delta$. We then categorize a ground-truth depth label $d$ into the $k$-th bin as:
\begin{align}
\small
\label{depth_category}
    k = \lfloor -0.5 + 0.5\sqrt{{1+\frac{8(d-d_{min})}{\delta}}} \rfloor,
\end{align} 
where $\delta = \frac{2(d_{max}-d_{min})}{k(k+1)}$.
By focusing on the object-wise depth values, the network can better capture foreground spatial structures and inter-object depth relations, which produces informative depth features for the subsequent depth-guided transformer.

\vspace{0.15cm}
\subsection{Depth-guided Transformer}
\label{s3.2}
The depth-guided transformer of MonoDETR is composed of a visual encoder, a depth encoder, and a depth-guided decoder. The two encoders produce non-local visual and depth embeddings, and the decoder enables object queries to adaptively capture scene-level information.
\vspace{-0.2cm}

\paragraph{Visual and Depth Encoders.} 
Given depth and visual features $f_D, f_V$, we specialize two transformer encoders to generate their scene-level embeddings with global receptive fields, denoted as $f^e_D \in \mathbb{R}^{\frac{HW}{16^2}\times C}$ and $f^e_V \in \mathbb{R}^{\frac{HW}{32^2}\times C}$.
We set three blocks for the visual encoder and only one block for the depth encoder, since the discrete foreground depth information is easier to be encoded than the rich visual appearances.
Each encoder block consists of a self-attention layer and a feed-forward neural network (FFN).
By the global self-attention mechanism, the depth encoder explores long-range dependencies of depth values from different foreground areas, which provides non-local geometric cues of the stereo space. In addition, the decoupling of depth and visual encoders allows them to better learn features for themselves, encoding the input image from two perspectives, i.e., depth geometry and visual appearance.

\begin{figure*}[t!]
  \centering
    \includegraphics[width=0.95\textwidth]{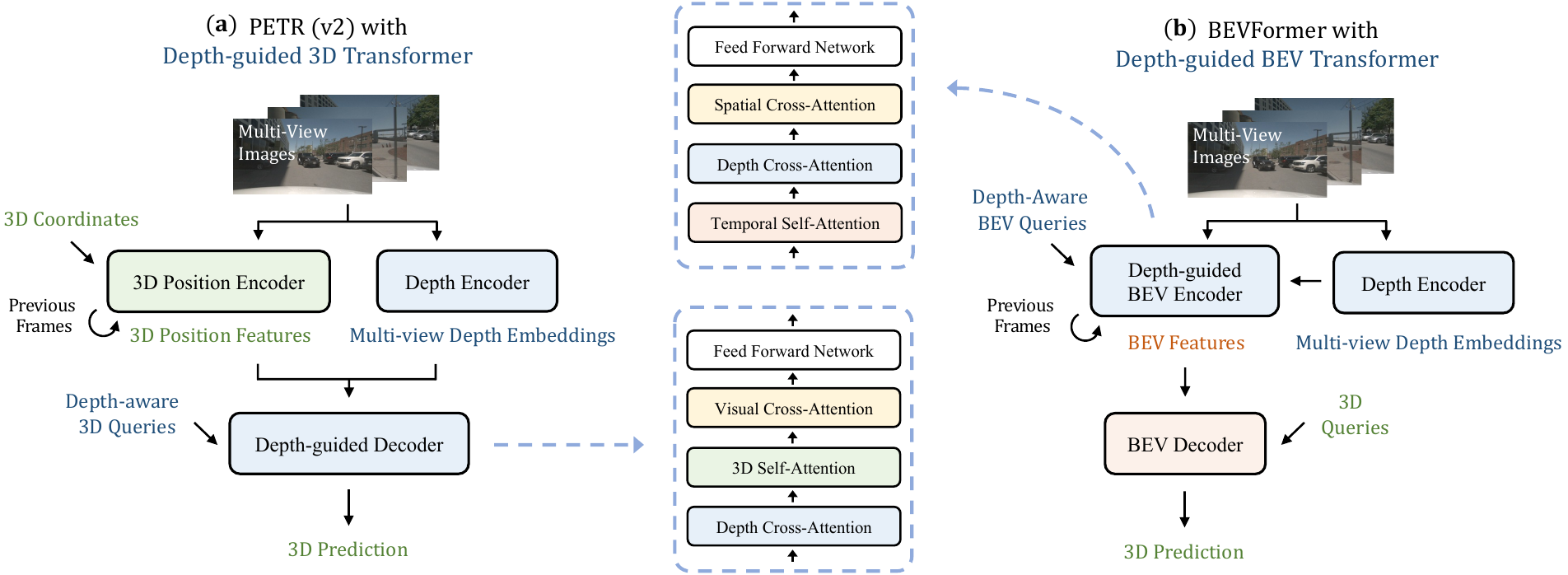}
    \vspace{0.2cm}
   \caption{\textbf{Plug-and-play for multi-view 3D object detection.} We utilize yellow, blue, green, and red to respectively denote the feature space related to 2D, depth, 3D, and BEV. The depth-guided transformer of MonoDETR is adopted to enhance PETR (v2)~\cite{liu2022petr,liu2022petrv2} and BEVFormer~\cite{li2022bevformer} in a plug-and-play manner, which provides depth guidance from surrounding scenes.}
    \label{fig5}
\end{figure*}

\paragraph{Depth-guided Decoder.}
Based on the non-local $f^e_D, f^e_V$, we utilize a set of learnable object queries $q\in \mathbb{R}^{N\times C}$ to detect 3D objects via the depth-guided decoder, where $N$ denotes the pre-defined maximum number of objects in the input image. Each decoder block sequentially contains a depth cross-attention layer, an inter-query self-attention layer, a visual cross-attention layer, and an FFN. 
Specifically, the queries first capture informative depth features from $f^e_D$ via the depth cross-attention layer, in which we linearly transform the object queries and depth embeddings into queries, keys, and values,
\begin{align}
\label{transform}
    Q_q = \mathrm{Linear}(q), \ \ \ \ K_D, V_D = \mathrm{Linear}(f_D^e),
\end{align}
where $Q_q\in \mathbb{R}^{N\times C}$ and $K_D, V_D \in \mathbb{R}^{\frac{HW}{16^2}\times C}$.
Then, we calculate the query-depth attention map $A_D \in \mathbb{R}^{N\times\frac{HW}{16^2}}$, and aggregate informative depth features weighted by $A_D$ to produce the depth-aware queries $q'$, formulated as,
\begin{align}
\label{depth_attention}
    A_D &= \mathrm{Softmax}(Q_qK_D^T/\sqrt{C}),\\
    q' &= \mathrm{Linear}(A_D V_D).
\end{align}
Such a mechanism enables each object query to adaptively capture spatial cues from depth-guided regions on the image, leading to better scene-level spatial understanding.
Then, the depth-aware queries are fed into the inter-query self-attention layer for feature interaction between objects, and the visual cross-attention layer for collecting visual semantics from $f_V^e$. We stack three decoder blocks to fully fuse the scene-level depth cues into object queries.

\paragraph{Depth Positional Encodings.} In the depth cross-attention layer, we propose learnable depth positional encodings for $f^e_D$ instead of conventional sinusoidal functions. In detail, we maintain a set of learnable embeddings, $p_D \in \mathbb{R}^{(d_{max}-d_{min}+1)\times C}$, where each row encodes the depth positional information for a meter, ranging from $d_{min}$ to $d_{max}$. For each pixel $(x, y)$ in $D_{fg}$, we first obtain its $(k+1)$-categorical depth prediction confidence, $D_{fg}(x, y)\in \mathbb{R}^{k+1}$, each channel of which denotes the predicted confidence for the corresponding depth bin. The estimated depth of pixel $(x, y)$ can then be obtained by the weighted summation of the depth-bin confidences and their corresponding depth values, which is formulated as
\begin{align}
\label{depth_sample}
    d_{map}(x, y) = \sum_{i=1}^{k+1} D_{fg}(x, y)[i] \cdot d_{bin}^i,
\end{align}
where $d_{bin}^i$ denotes the starting value of the $i$-th depth bin and $\sum_{i=1}^{k+1} D_{fg}(x, y)[i] = 1$.
Then, we linearly interpolate $p_D$ according to the depth $d_{map}(x, y)$ to obtain the depth positional encoding for the pixel $(x, y)$. By pixel-wisely adding $f^e_D$ with such encodings, object queries can better capture scene-level depth cues and understand 3D geometry in the depth cross-attention layer.

\subsection{Detection Heads and Loss}
\label{s3.3}
After the decoder, the depth-aware queries are fed into a series of MLP-based heads for 3D attribute predictions, including object category, 2D size, projected 3D center, depth, 3D size, and orientation. For inference, we convert these perspective attributes into 3D-space bounding boxes using camera parameters without NMS post-processing or pre-defined anchors. For training, we match the orderless queries with ground-truth labels and compute losses for the paired ones. We refer to Supplementary Material for details.
\vspace{-0.6cm}

\paragraph{Bipartite Matching.}
To correctly match each query with ground-truth objects, we calculate the loss for each query-label pair and utilize Hungarian algorithm~\cite{DETR} to find the globally optimal matching. For each pair, we integrate the losses of six attributes into two groups. The first contains object category, 2D size and the projected 3D center, since these attributes mainly concern 2D visual appearances of the image. 
The second group consists of depth, 3D size and orientation, which are 3D spatial properties of the object. We respectively sum the losses of two groups and denote them as $\mathcal{L}_{2D}$ and $\mathcal{L}_{3D}$. As the network generally predicts less accurate 3D attributes than 2D attributes, especially at the beginning of training, the value of $\mathcal{L}_{3D}$ is unstable and would disturb the matching process. We only utilize $\mathcal{L}_{2D}$ as the matching cost for matching each query-label pair.
\vspace{-0.3cm}

\paragraph{Overall Loss.}
After the matching, we obtain $N_{gt}$ valid pairs out of $N$ queries, where $N_{gt}$ denotes the number of ground-truth objects. Then, the overall loss of a training image is formulated as
\begin{align}
\label{all_loss}
    \mathcal{L}_{overall} = \frac{1}{N_{gt}}\cdot\sum_{n=1}^{N_{gt}}(\mathcal{L}_{2D} + \mathcal{L}_{3D}) + \mathcal{L}_{dmap},
\end{align}
where $\mathcal{L}_{dmap}$ represents the Focal loss~\cite{lin2017focal} of the predicted categorical foreground depth map $D_{fg}$ in Section~\ref{s3.1}. 

\subsection{Plug-and-play for Multi-view Detectors}
\label{s3.4}
Besides monocular images, our depth-guided transformer can also serve as a plug-and-play module upon multi-view methods for depth-guided detection. Specifically, we append our depth predictors and depth encoders after the backbones of multi-view methods, which are shared across views and extract surrounding depth embeddings. Then, we inject our depth cross-attention layer into their transformer blocks to guide the 3D or BEV object queries by scene-level depth cues. 
\vspace{-0.3cm}

\begin{table*}[t!]
\centering
\small
\begin{tabular}{l|c|ccc|ccc|ccc}
	\toprule
\multirow{2}{*}{Method} & \multirow{2}{*}{Extra data} & \multicolumn{3}{c|}{Test,\ $AP_{3D}$} & \multicolumn{3}{c|}{Test,\ $AP_{BEV}$} & \multicolumn{3}{c}{Val,\ $AP_{3D}$} \\ 
& & Easy & Mod. & Hard & Easy & Mod. & Hard & Easy & Mod. & Hard \\
\midrule
PatchNet~\cite{patchnet} & \multirow{3}{*}{Depth}   & 15.68 & 11.12 & 10.17 & 22.97 & 16.86 & 14.97 & -  & -  & - \\ 
D4LCN~\cite{D4LCN} &             & 16.65 & 11.72 & 9.51  & 22.51 & 16.02 & 12.55 & - & - & - \\
DDMP-3D~\cite{ddpm} &                     & 19.71 & 12.78 & 9.80  & 28.08 & 17.89 & 13.44 & - & - & - \\
\midrule
Kinematic3D~\cite{Kinematic3D}  & Video      & 19.07 & 12.72 & 9.17  & 26.69 & 17.52 & 13.10 & 19.76 & 14.10 & 10.47 \\
\midrule
MonoRUn~\cite{monorun}  & \multirow{3}{*}{LiDAR}    & 19.65 & 12.30 & 10.58 & 27.94 & 17.34 & 15.24 & 20.02 & 14.65 & 12.61 \\
CaDDN~\cite{CaDDN} &                                & 19.17 & 13.41 & 11.46 & 27.94 & 18.91 & \color{blue}{17.19} & 23.57 & 16.31 & 13.84 \\
MonoDTR~\cite{monodtr} &                                & 21.99 & \color{blue}{15.39} & \color{blue}{12.73} & 28.59 & \color{blue}{20.38} & 17.14 & \color{blue}{24.52} & \color{blue}{18.57} & 15.51\\
\midrule
AutoShape~\cite{AutoShape} &  CAD             & \color{blue}{22.47} & 14.17 & 11.36 & \color{blue}{30.66} & 20.08 & 15.59 & 20.09 & 14.65 & 12.07 \\
\midrule
SMOKE~\cite{smoke} &     \multirow{10}{*}{None}                           & 14.03 & 9.76  & 7.84  & 20.83 & 14.49 & 12.75 & 14.76 & 12.85 & 11.50 \\
MonoPair~\cite{MonoPair} &                          & 13.04 & 9.99  & 8.65  & 19.28 & 14.83 & 12.89 & 16.28 & 12.30 & 10.42\\
RTM3D~\cite{RTM3D} &         & 13.61 & 10.09 & 8.18  & -     & -     & -     & 19.47 & 16.29 & \color{blue}{15.57} \\
PGD~\cite{PGD} &                          & 19.05 & 11.76 & 9.39 & 26.89 & 16.51 & 13.49 & 19.27 & 13.23 & 10.65 \\ 
IAFA~\cite{iafa} &                                  & 17.81 & 12.01 & 10.61 & 25.88 & 17.88 & 15.35 & 18.95 & 14.96 & 14.84 \\ 
MonoDLE~\cite{monodle} &                            & 17.23 & 12.26 & 10.29 & 24.79 & 18.89 & 16.00 & 17.45 & 13.66 & 11.68 \\ 
MonoRCNN~\cite{monorcnn} &                          & 18.36 & 12.65 & 10.03 & 25.48 & 18.11 & 14.10 & 16.61 & 13.19 & 10.65 \\ 
MonoGeo~\cite{monogeo} &                            & 18.85 & 13.81 & 11.52 & 25.86 & 18.99 & 16.19 & 18.45 & 14.48 & 12.87 \\ 
MonoFlex~\cite{MonoFlex} &                          & 19.94 & 13.89 & 12.07 & 28.23 & 19.75 & 16.89 & 23.64 & 17.51 & 14.83 \\ 
GUPNet~\cite{gupnet} &                              & 20.11 & 14.20 & 11.77 & -     & -     & -     & 22.76 & 16.46 & 13.72 \\ 
\midrule
\textbf{MonoDETR~(Ours)} & None & \textbf{25.00} & \textbf{16.47} & \textbf{13.58} & \textbf{33.60} & \textbf{22.11} & \textbf{18.60}  &\textbf{28.84} &\textbf{20.61} &\textbf{16.38} \vspace{0.1cm}\\
\textit{Improvement} &\textit{v.s. second-best} &\color{blue}{+2.53} &\color{blue}{+1.08} &\color{blue}{+0.85} &\color{blue}{+2.94} &\color{blue}{+1.73} &\color{blue}{+1.41} &\color{blue}{+4.32} &\color{blue}{+2.04} &\color{blue}{+0.81}\\
\bottomrule
\end{tabular}
\vspace{0.2cm}
\caption{\textbf{Monocular performance of the car category on KITTI \textit{test} and \textit{val} sets.} We utilize bold numbers to highlight the best results, and color the second-best ones and our gain over them in blue.}
\label{perf}
\end{table*}


\paragraph{For PETR (v2)~\cite{liu2022petr,liu2022petrv2}} in Figure~\ref{fig5} (a), we modify its previous visual decoder as a depth-guided decoder. In each decoder block, the 3D object queries are first fed into our depth cross-attention layer for depth cues aggregation, and then into the original 3D self-attention and visual cross-attention for 3D position features interaction. This enables PETR's 3D queries to be depth-aware and better capture spatial characteristics of surrounding scenes.
\vspace{-0.3cm}

\paragraph{For BEVFormer~\cite{li2022bevformer}} in Figure~\ref{fig5} (b), as its decoder is conducted in BEV space, we incorporate the depth guidance into its BEV encoder, which lifts image features into BEV space by transformers. In each encoder block, the BEV queries also sequentially pass through our depth cross-attention layer and the original spatial cross-attention layer. This contributes to better BEV representation learning guided by the multi-view depth information.

\section{Experiments}
\label{sec:experiments}

\subsection{Settings}

\paragraph{Dataset.} We evaluate MonoDETR on the widely-adopted KITTI~\cite{kitti} benchmark, including 7,481 training and 7,518 test images. We follow~\cite{Mono3D,3DOP} to split 3,769 \textit{val} images from the training set. 
We report the detection results with three-level difficulties, easy, moderate, and hard, and evaluate by the average precision ($AP$) of bounding boxes in 3D space and the bird-eye view, denoted as $AP_{3D}$ and $AP_{BEV}$, respectively, which are both at 40 recall positions. 

\vspace{-0.25cm}

\paragraph{Implementation details.}
We adopt ResNet-50~\cite{resnet} as our feature backbone. To save GPU memory, we apply deformable attention~\cite{zhu2020deformable} for the visual encoder and visual cross-attention layers, and utilize the vanilla global attention~\cite{DETR} to better capture non-local geometries for the depth encoder and depth cross-attention layers. We utilize 8 heads for all attention modules and set the number of queries $N$ as 50, which are learnable embeddings with predicted 2D reference points. We set the channel $C$ and all MLP's latent dimensions as 256. For the foreground depth map, we set $[d_{min}, d_{max}]$ as $[0m, 60m]$ and the number of bins $k$ as 80. On a single RTX 3090 GPU, we train MonoDETR for 195 epochs with batch size 16 and a learning rate $2\times 10^{-4}$. We adopt AdamW~\cite{adamW} optimizer with weight decay $10^{-4}$, and decrease the learning rate at 125 and 165 epochs by 0.1.  
For training stability, we discard the training samples with depth labels larger than 65 meters or smaller than 2 meters. 
During inference, we simply filter out the object queries with the category confidence lower than 0.2 without NMS post-processing, and recover the 3D bounding box using the predicted six attributes following previous works~\cite{monodle,gupnet}.

\vspace{0.1cm}
\subsection{Comparison}
\vspace{0.1cm}

\paragraph{Performance.} In Table~\ref{perf}, MonoDETR achieves \textit{state-of-the-art} performance on KITTI \textit{test} and \textit{val} sets.
On \textit{test} set, MonoDETR exceeds all existing methods including those with different additional data input and surpasses the second-best under three-level difficulties by +2.53\%, +1.08\% and +0.85\% in $AP_{3D}$, and by +2.94\%, +1.73\% and +1.41\% in $AP_{BEV}$.
The competitive MonoDTR~\cite{monodtr} also applies transformers to fuse depth features, but it is still a center-guided method and highly relies on additional dense depth supervision, anchors and NMS.
In contrast, MonoDETR performs better without extra input or handcrafted designs, illustrating its simplicity and effectiveness.
\vspace{-0.2cm}

\paragraph{Efficiency.}
Compared to existing methods in Table~\ref{t2}, MonoDETR can achieve the best detection performance without consuming too much computational budget. As illustrated in Section~\ref{sec:method}, we only process the feature maps with $\frac{1}{16}$ and $\frac{1}{32}$ downsample ratios, which reduces our Runtime and GFlops, while others adopt $\frac{1}{4}$ and $\frac{1}{8}$ ratios.
\vspace{0.1cm}

\subsection{Ablation Studies}

We verify the effectiveness of each our component and report $AP_{3D}$ for the car category on the KITTI \textit{val} set.
\vspace{-0.1cm}

\paragraph{Depth-guided Transformer.} 
In Table~\ref{t5}, we first remove the entire depth-guided transformer along with the depth predictor, which constructs a pure center-guided baseline. This variant, denoted as `w/o Depth-guided Trans.' can be regarded as a re-implementation of MonoDLE~\cite{monodle} with our detection heads and loss functions. As shown, the absence of the depth-guided transformer greatly hurts the performance, for the lack of non-local geometric cues. Then, we investigate two key designs within the depth-guided transformer: the transformer architecture and depth guidance. For `w/o Transformer', we only append the depth predictor upon the center-guided baseline to provide implicit depth guidance without transformers. For `w/o Depth Guidance', we equip the center-guided baseline with a visual encoder and decoder, but include no depth predictor, depth encoder, and the depth cross-attention layer in the decoder. This builds a transformer network guided by visual appearances, without any depth guidance for object queries.
The performance degradation of both variants indicates their significance for our depth-guided feature aggregation paradigm. 
\vspace{-0.2cm}

\begin{table}[t]
\centering
\begin{adjustbox}{width=\linewidth}
	\begin{tabular}{l c c c c }
	\toprule
	\multirow{1}{*}{Method} &MonoDLE &GUPNet &MonoDTR &MonoDETR\\
	\cmidrule(lr){1-1} \cmidrule(lr){2-2} \cmidrule(lr){3-3} \cmidrule(lr){4-4} \cmidrule(lr){5-5}
	Runtime$\downarrow$ &40 &\textbf{34} &37 &38\\
        GFlops$\downarrow$ &79.12 &62.32 &120.48 &\textbf{62.12}\\
        $AP_{3D}$ Mod. &12.26 &15.02 &15.39 &\textbf{16.47}\\
	\bottomrule
	\end{tabular}
\end{adjustbox}
\vspace{0.05cm}
\caption{\textbf{Efficiency comparison.} We test the Runtime (ms) on one RTX 3090 GPU with batch size 1, and compare $AP_{3D}$ on \textit{test} set.}
\label{t2}
\end{table}

\begin{table*}[t!]
\centering
\small
\begin{tabular}{l|c|cc|ccccc}
\toprule
Method & Image Size &NDS$\uparrow$ &mAP$\uparrow$ &mATE$\downarrow$ &mASE$\downarrow$ &mAOE$\downarrow$ &mAVE$\downarrow$ &mAAE$\downarrow$ \\
\midrule
CenterNet~\cite{CenterNet} &- &0.328 &0.306 &0.716 &0.264 &0.609 &1.426 &0.658\\
FCOS3D*~\cite{fcos3d} &1600$\times$900 &0.415 &0.343 &0.725 &0.263 &0.422 &1.292 &0.153\\
PGD*~\cite{PGD} &1600$\times$900 &0.428 &0.369 &0.683 &0.260 &0.439 &1.268 &0.185\\
DETR3D$\dagger$~\cite{detr3d} &1600$\times$900 &0.434 &0.349 &0.716 &0.268 &0.379 &0.842 &0.200\\
BEVDet$\dagger$~\cite{huang2021bevdet} &1408$\times$512 &0.417 &0.349 &0.637 &0.269 &0.490 &0.914 &0.268\\
PETR$\dagger$~\cite{liu2022petr} &1600$\times$900 &0.442 &0.370 &0.711 &0.267 &0.383 &0.865 &0.201\\
\midrule
PETRv2~\cite{liu2022petrv2} &\multirow{2}{*}{800$\times$320} &0.496 &0.401 &0.745 &0.268 &0.448 &0.394 &0.184\\
\textit{\ \ + Depth-gudied} & &\textbf{0.508} &\textbf{0.410} &0.727 &0.265 &0.389 &0.419 &0.187\\\midrule
BEVFormer~\cite{li2022bevformer} &\multirow{2}{*}{1600$\times$900} &0.517 &0.416 &0.673 &0.274 &0.372 &0.394 &0.198\\
\textit{\ \ + Depth-gudied} & &\textbf{0.526} &\textbf{0.423} &0.661 &0.272 &0.349 &0.371 &0.192\\
\bottomrule
\end{tabular}
\vspace{0.2cm}
\caption{\textbf{Multi-view performance on nuScenes \textit{val} set.} * denotes the two-step fine-tuning with test-time augmentation, and $\dagger$ denotes CBGS~\cite{CBGS} training. We compare with the best-performing variants of other methods and utilize bold numbers to highlight the best results.}
\label{nus}
\end{table*}

\begin{table}[]
\centering
\small
\begin{tabular}{c|ccc}
\toprule
Architecture & Easy & Mod. & Hard \\
\midrule
MonoDETR &\textbf{28.84} & \textbf{20.61} & \textbf{16.38}\vspace{0.1cm}\\
w/o Depth-guided Trans. & 19.69 & 15.15 & 13.93 \\
w/o Transformer & 20.19 & 16.05& 14.18 \\
w/o Depth Guidance & 24.14 & 17.81 & 15.60 \\
\bottomrule
\end{tabular}
\vspace{0.2cm}
\caption{\textbf{Effectiveness of depth-guided transformer.} 
`Depth-guided Trans.' and `Depth Guidance' denote the depth-guided transformer and the depth cross-attention layer, respectively}
\label{t5}
\end{table}

\paragraph{Depth Encoder.}
The depth encoder produces non-local depth embeddings $f^e_D$, which are essential for queries to explore scene-level depth cues in the depth cross-attention layer. We experiment with different encoder designs in Table~\ref{t6}. `Deform. SA' and `3$\times$3 Conv.$_{\times2}$' represent one-block of deformable attention and two $3\times3$ convolutional layers, respectively.
As reported, `Global SA' with only one block generates the best $f^e_D$ for global gemoetry encoding.
\vspace{-0.3cm}

\paragraph{Depth-guided Decoder.}
As the core depth-guided component, we explore how to better guide object queries to interact with depth embeddings $f^e_D$ in Table~\ref{t7}. With the sequential inter-query self-attention (`$I$') and visual cross-attention (`$V$') layers, we insert the depth cross-attention layer (`$D$') into each decoder block with four positions. For `$I\rightarrow D + V$', we fuse the depth and visual embeddings $f_D^e, f_V^e$ by element-wise addition, and apply only one unified cross-attention layer. As shown, the `$D\rightarrow I\rightarrow V$' order performs the best. By placing `$D$' in the front, object queries can first aggregate depth cues to guide the remaining operations in each decoder block.
\vspace{-0.3cm}

\paragraph{Foreground Depth Map.}
We explore different representations for our depth map in Table~\ref{t8}. Compared to dense depth supervision (`Dense'), adopting object-wise depth labels (`Fore.') can focus the network on more important foreground geometric cues, and better capture depth relations between objects.
`LID' outperforms other discretization methods, since the linear-increasing intervals can suppress the larger estimation errors of farther objects.

\paragraph{Depth Positional Encodings $p_D$.}
In Table~\ref{depthpos}, we experiment with different depth positional encodings for $f_D^e$ in the depth cross-attention layer. By default, we apply the meter-wise encodings $p_D \in \mathbb{R}^{(d_{max}-d_{min}+1)\times C}$ that assign one learnable embedding per meter with depth value interpolation for output. We then assign one learnable embedding for each depth bin, denoted as `$k$-bin $p_D \in\mathbb{R}^{k\times C}$', and also experiment sinusoidal functions to encode either the depth values or 2D coordinates of the feature map, denoted as `Depth sin/cos' and `2D sin/cos', respectively. As shown, `meter-wise $p_D$' performs the best for encoding more fine-grained depth cues ranging from $d_{min}$ to $d_{max}$, which provides the queries with more scene-level spatial structures.

\begin{table}[t]
\centering
\small
\begin{tabular}{c|ccc}
\toprule
Mechanism & Easy & Mod. & Hard \\
\midrule
Global SA &\textbf{28.84} & \textbf{20.61} & \textbf{16.38}\vspace{0.1cm}\\
Deform. SA & 26.43 & 18.91 & 15.55 \\
3$\times$3 Conv.$_{\times2}$ & 25.55 & 18.36 & 15.28 \\
w/o & 24.25 & 18.38 & 15.41 \\
\bottomrule
\end{tabular}
\vspace{0.2cm}
\caption{\textbf{The design of depth encoder.} 
`Deform.\hspace{-0.05cm} SA' denotes a one-block deformable self-attention layer. `w/o' denotes directly feeding depth features into the decoder without the depth encoder.}
\label{t6}
\end{table}

 \begin{figure*}[t!]
  \centering
    \includegraphics[width=\textwidth]{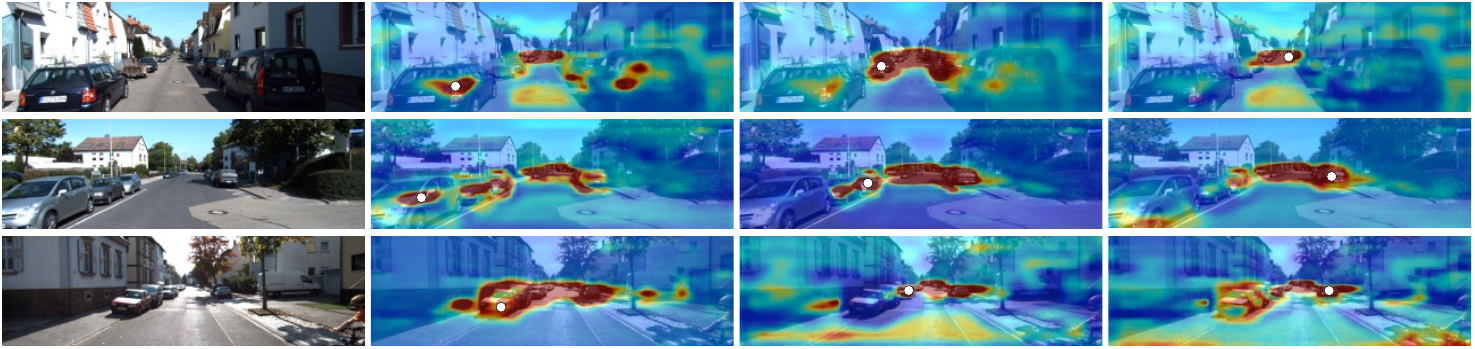}
   \caption{Visualizations of attention maps $A_D$ in the depth cross-attention layer. The top column denotes the input image, and the last three columns denote the attention maps of the target queries (denoted as white dots). Hotter colors indicate higher attention weights.}
    \label{fig:activation}
\end{figure*}

\subsection{Multi-view Experiments}

As a plug-and-play module for multi-view 3D object detection, we append our depth-guided transformer upon two DETR-based multi-view networks, PETR v2~\cite{liu2022petrv2} and BEVFormer~\cite{li2022bevformer}. The detailed network architectures are shown in Figure~\ref{t5}. For a fair comparison, we adopt the same training configurations as the two baseline models, and utilize the same $\left[d_{min}, d_{max}\right]$ with $k$ as monocular experiments. We report the performance on nuScenes~\cite{nuScenes} \textit{val} set in Table~\ref{nus}, where we apply no test-time augmentation or CGBS~\cite{CBGS} training. For end-to-end detection in PETRv2, the depth guidance contributes to +1.2\% NDS and +0.9\% mAP by providing sufficient multi-view geometric cues. For the BEV feature generation, our modules benefit BEVFormer by +0.9\% NDS and +0.7\% mAP, indicating the importance of auxiliary depth information for BEV-space feature encoding. The additional experiments on multi-view 3D object detection well demonstrate the effectiveness and generalizability of our approach.


\vspace{0.1cm}
\section{Visualization}
We visualize the attention maps $A_D$ in Equation~\ref{depth_attention} of the depth cross-attention layer at the last decoder block. As shown in Figure~\ref{fig:activation}, the areas with high attention scores for the target object query spread over the entire image, concentrating on other objects with long distances. This indicates, via our depth guidance, object queries are able to adaptively capture non-local depth cues from the image and are no longer limited by neighboring visual features.

\begin{table}[t]
\centering
\small
\begin{tabular}{c|ccc}
\toprule
\ \ \ \ \ \ \ Architecture\ \ \ \ \ \  & Easy & Mod. & Hard \\
\midrule
D $\rightarrow$ I$\rightarrow$ V &\textbf{28.84} & \textbf{20.61} & \textbf{16.38}\vspace{0.1cm}\\
I $\rightarrow$ D$\rightarrow$ V & 26.24 & 19.28 & 16.03 \\
I $\rightarrow$ V$\rightarrow$ D & 25.84 & 18.85 & 15.72 \\
I\ $\rightarrow$\ \ D + V & 24.94 & 18.41 & 15.39 \\
\bottomrule
\end{tabular}
\vspace{0.25cm}
\caption{\textbf{The design of depth-guided decoder.} `D', `I', and `V' denote the depth cross-attention, inter-query self-attention, and visual cross-attention layers, respectively.}
\label{t7}
\end{table}

\vspace{0.1cm}
\section{Conclusion}
\label{sec:conclusion}
We propose MonoDETR, an end-to-end transformer-based framework for monocular 3D object detection, which is free from any additional input, anchors, or NMS. Different from existing center-guided methods, we enable object queries to explore geometric cues adaptively from the depth-guided regions, and conduct inter-object and object-scene depth interactions via attention mechanisms.
Extensive experiments have demonstrated the effectiveness of our approach for both single-view (KITTI) and multi-view (nuScenes) input. We hope MonoDETR can serve as a strong DETR baseline for future research in monocular 3D object detection. \textbf{Limitations.} How to effectively incorporate multi-modal input into our transformer framework is not discussed in the paper. Our future direction will focus on this to further improve the performance of depth-guided transformers, e.g., distilling more sufficient geometric knowledge from LiDAR and RADAR modalities.

\section*{Acknowledgement}
This project is funded in part by the National Natural Science Foundation of China (No.62206272), by the National Key R\&D Program of China Project (No.2022ZD0161100), by the Centre for Perceptual and Interactive Intelligence (CPII) Ltd under the Innovation and Technology Commission (ITC)’s InnoHK, and by the General Research Fund of Hong Kong RGC Project 14204021. Hongsheng Li is a PI of CPII under the InnoHK.

\begin{table}[t]
\centering
\small
\begin{tabular}{c|ccc}
\toprule
\ \ \ \ \ \ Depth Map \ \ \ \ \ \ & Easy & Mod. & Hard \\
\midrule
Fore. LID &\textbf{28.84} & \textbf{20.61} & \textbf{16.38}\vspace{0.15cm}\\
Dense LID &27.69 &19.85 &15.98 \\
Fore. UD & 25.61 & 18.90 & 15.49 \\
Fore. SID & 26.05 & 18.95 & 15.59 \\
\bottomrule
\end{tabular}
\vspace{0.2cm}
\caption{\textbf{Different representations of the predicted depth map.} `UD', `SID', and `LID' denote uniform, spacing-increasing, and linear-increasing discretizations.}
\label{t8}
\vspace{0.3cm}
\end{table}

\begin{table}[]
\centering
\small
\begin{tabular}{l|ccc}
\toprule
Settings & Easy & Mod. & Hard \\
\midrule
Meter-wise $p_D$ &\textbf{28.84} & \textbf{20.61} & \textbf{16.38}\vspace{0.1cm}\\
$k$-bin $p_D$ & 28.06 & 19.68 & 16.04 \\
Depth sin/cos & 27.42 & 19.57 & 15.82 \\
2D sin/cos & 26.48 & 18.63 & 15.52 \\
w/o & 26.76 & 18.94 & 15.85 \\
\bottomrule
\end{tabular}
\vspace{0.2cm}
\caption{\textbf{The design of depth positional encodings.} `Meter-wise' and `$k$-bin' assign learnable embeddings by meters and depth bins, respectively. `sin/cos' denotes sinusoidal functions for encodings.}
\label{depthpos}
\end{table}


{\small
\bibliographystyle{ieee_fullname}
\bibliography{egbib}
}

\end{document}